\def\year{2019}\relax
\documentclass[letterpaper]{article} %DO NOT CHANGE THIS
\usepackage{formatting-instructions-latex-2019}  %Required
\usepackage{times}  %Required
\usepackage{helvet}  %Required
\usepackage{courier}  %Required
\usepackage{url}  %Required
\usepackage{graphicx}  %Required
\usepackage{amsmath,amssymb} % define this before the line numbering.
\usepackage{bm}
\usepackage{booktabs}
\usepackage{multirow}
\usepackage{epstopdf}
\usepackage[ruled]{algorithm2e}
\usepackage{amsmath}
\usepackage{subfigure}

\newcommand{\eg}{\emph{e.g. }}
\newcommand{\ie}{\emph{i.e. }}

\newcommand{\tabincell}[2]{\begin{tabular}{@{}#1@{}}#2\end{tabular}}

\begin{document}
\title{Devil in the Details: Towards Accurate Single and Multiple Human Parsing}  

\author{Tao~Ruan$^{1*}$, Ting~Liu$^{1*}$, Zilong~Huang$^{3}$, Yunchao~Wei$^{2\dagger}$, Shikui~Wei$^{1}$, Yao~Zhao$^{1}$, Thomas~Huang$^{2}$ \\
$^{1}$Institute of Information Science, Beijing Jiaotong University, China\\
$^{2}$University of Illinois at Urbana-Champaign, USA\\
$^{3}$School of Electronic Information and Communications, Huazhong University of Science and Technology, China\\
}

%\author{Ting~Liu\footnote[1]{ddd}, Tao~Ruan\footnote[1]{ddd}, Zilong~Huang, Yunchao Wei, Shikui~Wei, Yao~Zhao,~\IEEEmembership{Senior Member,~IEEE,} Thomas~Huang~\IEEEmembership{Life~Fellow,~IEEE}

\maketitle
 \begin{abstract}

Human parsing has received considerable interest due to its wide application potentials. Nevertheless, it is still unclear how to develop an accurate human parsing system in an efficient and elegant way. In this paper, we identify several useful properties, including feature resolution, global context information and edge details, and perform rigorous analyses to reveal how to leverage them to benefit the human parsing task. The advantages of these useful properties finally result in a simple yet effective Context Embedding with Edge Perceiving (CE2P) framework for single human parsing. Our CE2P is end-to-end trainable and can be easily adopted for conducting multiple human parsing. Benefiting the superiority of CE2P, we won the 1st places on all three human parsing tracks in the 2nd Look into Person (LIP) Challenge. Without any bells and whistles, we achieved 56.50\% (mIoU), 45.31\% (mean $AP^r$) and 33.34\% ($AP^p_{0.5}$) in Track 1, Track 2 and Track 5, which outperform the state-of-the-arts more than 2.06\%, 3.81\% and 1.87\%, respectively. We hope our CE2P will serve as a solid baseline and help ease future research in single/multiple human parsing. Code has been made available at \url{https://github.com/liutinglt/CE2P}.

 \end{abstract}

\section{Introduction}

Human parsing is a fine-grained semantic segmentation task which aims at identifying the constituent parts (\eg body parts and clothing items) for a human image on pixel-level. Understanding the contents of the human image makes sense in several potential applications including e-commerce, human-machine interaction, image editing and virtual reality to name a few. Currently, human parsing has gained remarkable improvement with the development of fully convolutional neural networks on semantic segmentation. 

For the semantic segmentation, researchers have developed many solutions from different views to tackle the challenging dense prediction task. In general, current solutions can be grossly divided into two types: 1) \textbf{High-resolution Maintenance} This kind of approaches are attempting to obtain high-resolution features to recover the desired detailed information. Due to consecutive spatial pooling and convolution strides, the resolution of the final feature maps is reduced significantly that the finer image information is lost. To generate high-resolution features, there are two typical solutions, \ie cutting several down-sampling (\eg max pooling) operations~\cite{Chen2016} and introducing details from low-level feature maps~\cite{Chen2018a}. For the latter case, it is usually embedded in an encoder-decoder architecture, in which the high-level semantic information is captured in the encoder and the details and spatial information are recovered in the decoder. 2) \textbf{Context Information Embedding} This kind of approaches is devoting to capture rich context information to handle the objects with multiple scales. Feature pyramid is one of the effective ways to mitigate the problem caused by various scales of objects, and atrous spatial pyramid pooling (ASPP) based on dilated convolution~\cite{Chen2018} and pyramid scene parsing (PSP)~\cite{Zhao2017} are two popular structures. ASPP utilizes parallel dilated convolution layers with different rates to incorporate multi-scale context. PSP designs pyramid pooling operation to integrate the local and global information together for more reliable prediction. Beyond these two typical types, some other works also propose to advantage the segmentation performance by introducing additional information such as edge~\cite{Liang2017,liumagic} or more effective learning strategy such as cascaded training~\cite{Li_2017_CVPR}.

%dilated (or atrous) convolution and encoder-decoder architecture. The dilated convolution is applied to enlarge the receptive fields so that we can use pre-trained networks with the downsampling operations from last few layers removed. In this manner, we can decide the resolution of features without extra parameters. The encoder-decoder architecture includes two parts, which named encoder and decoder. Specifically, the high-level semantic information is captured in the encoder, and the details and spatial information are recovered in the decoder. One way used in decoder architecture is applying deconvolution or unpooling as up-sampling operation to produce dense feature responses. The other way is exploiting intermediate layers with low-level spatial visual information as complementary information for deeper layers.  2) The second one is devoting to capture rich context information to handle the objects with multiple scales. Feature pyramid is one of the effective ways to mitigate the problem caused by various scales of objects, and atrous spatial pyramid pooling (ASPP) and pyramid scene parsing (PSP) are two typical structures. ASPP utilizes parallel dilated convolution layers with different rates to incorporate multi-scale context. PSP designs pyramid pooling operation to integrate the local and global information together for more reliable prediction.

Compare with the general semantic segmentation tasks, the challenges of human parsing is to produce finer predictions for every detailed region belonging to a person. Besides, the arms, legs and shoes are further divided into the left side and right side for more precise analysis, which makes the parsing more difficult. Despite the approaches mentioned above showing impressive results in semantic segmentation, it remains unclear how to develop an accurate human parsing system upon these solutions and most previous works did not explore and analyze how to leverage them to unleash the full potential in human parsing. In this work, we target on answering such a question: can we simply formulate a powerful framework for human parsing by exploiting the recent advantages in the semantic segmentation area?
  
%The difficulty of human parsing is generating finer predictions for every detailed region of a person. Besides, the arms, legs and shoes are further divided into the left side and right side for more precise analysis, which makes the parsing more difficult. Despite several methods mentioned above showing impressive results in semantic segmentation, it remains unclear how to develop an accurate human parsing system using those methods. However, most papers did not explore and analyze how to leverage them to unleash the full potential in human parsing.  

To answer such a question, we conduct a great deal of rigorous experiments to clarify the key factors affecting the performance of human parsing. In particular, we perform an analysis of potential gains in mIoU score with different properties. The evaluated useful properties include feature resolution, context information and edge details. Based on the analysis, we present a simple yet effective Context Embedding with Edge Perceiving (CE2P) framework for single human parsing. The proposed CE2P consists of three key modules to learn for parsing in an end-to-end manner: 1) A high-resolution embedding module used to enlarge the feature map for recovering the details; 2) A global context embedding module used for encoding the multi-scale context information; 3) An edge perceiving module used to integrate the characteristic of object contour to refine the boundaries of parsing prediction. Our approach achieves state-of-the-art performance on all three human parsing benchmarks. Those results manifest that the proposed CE2P can provide consistent improvements over various human parsing tasks. The main contributions of our work are as follows:

\begin{itemize}

\item We analyze the effectiveness of several properties for human parsing, and reveal how to leverage them to benefit the human parsing task.

\item We design a simple yet effective CE2P framework by leveraging the useful properties to conduct human parsing in a simple and efficient way. 

\item Our CE2P brings a significant performance boost to all three human parsing benchmarks, outperforming the current state-of-the-art method by a large margin. 

\item Our code is available, which can serve as a solid baseline for the future research in single/multiple human parsing.

\end{itemize}

\section{Related Work}

\subsection{Human Parsing}
The study of human parsing has drawn more and more attention due to the wide range of potential application. The early works~\cite{Yamaguchi2012,Dong2013,Simo-Serra2014,Ladicky2013} performed the parsing with CRF framework and utilized the human pose estimation to assist the parsing.  A Co-CNN~\cite{Liang2017} architecture was proposed to hierarchically integrate the local and global context information and improved the performance greatly. Recently, SSL~\cite{Gong2017} introduced a self-supervised structure-sensitive loss, which was used for enforcing the consistency between parsing results and the human joint structures, to assist the parsing task. Following previous work, JPPNet~\cite{Liang2018} incorporated the human parsing and pose estimation task into a unified network. With multi-scale feature connections and iterative refinement, the parsing and pose tasks boosted each other simultaneously. Considering the practical application, several current works~\cite{Li2017a,Li2017,Zhao2018} focus on handling the scenario with multiple persons. Usually, it consisted of three sequential steps: object detection~\cite{he2017mask}, object segmentation and part segmentation. Besides, many research efforts~\cite{Girshick2014,Wang2015,Chen2016a,Hariharan2015} have been devoted into the object parts segmentation, which was similar to human parsing. Most of those works leveraged the multi-scale features to enhance the performance.
  
\subsection{Semantic Segmentation}

Human parsing is a fine-grained semantic segmentation task. Hence, the methods used in human parsing is similar to semantic segmentation. Since the fully convolutional network (FCN)~\cite{Long2015} has shown numerous improvements in semantic segmentation, many researchers~\cite{Chen2016,Jegou2017,wei2018revisiting,wei2017object,wei2017stc} have made efforts based on the FCN. Several pieces of work~\cite{Badrinarayanan2017,Ronneberger2015,Lin2017}, leveraged an encoder-decoder architecture with skip connection to recover the dense feature responses. Another literatures~\cite{Chen2016,YuKoltun2016,Chen2018a} exploited the dilated convolution for higher resolution output. Besides, the local and global information are integrated for generating more reliable prediction, such as~\cite{Chen2018,Zhao2017}.

\section{Context Embedding with Edge Perceiving}

\begin{figure*}[!htb]
    \centering
    \includegraphics[width=1\textwidth, height=80mm]{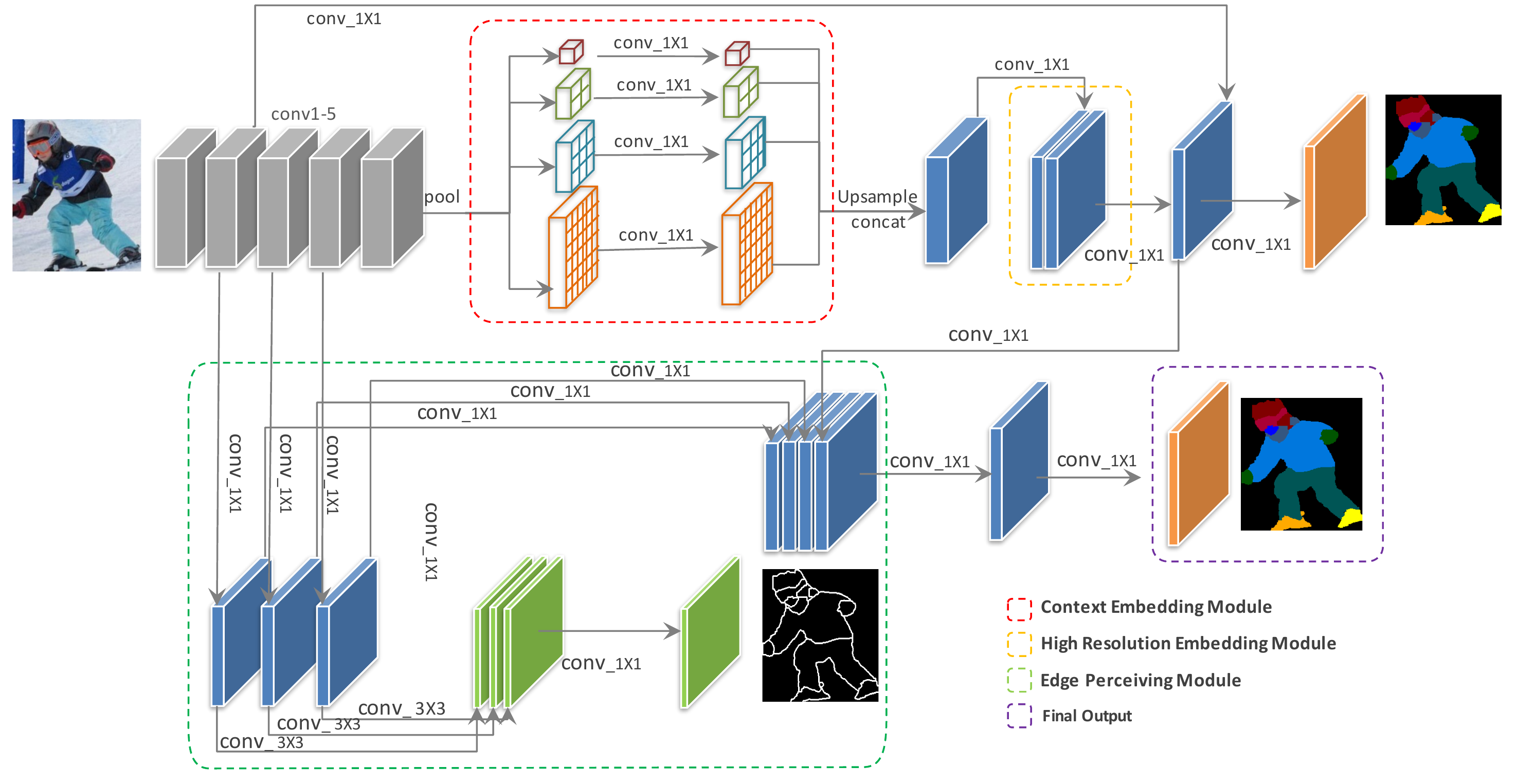}
    \caption{The framework of our proposed CE2P. The overall architecture consists of three key modules: 1) high resolution embedding module 2) global context embedding module; 3) edge perceiving module.  }
    \label{fig:framework}
\end{figure*}

In this section, we first provide the architecture of our Context Embedding with Edge Perceiving(CE2P) approach. Within CE2P, we analyze the effectiveness of several key modules motivated from the previous state-of-the-art semantic segmentation models and reveal how they can work together to accomplish the single human parsing task. Then, we give the details of applying the CE2P to address the more challenging multiple human parsing task.

%In this section, we first describe the architecture of our Context Embedding with Edge Perceiving(CE2P) in detail. In addition, we explain the approach used in multiple human parsing.
 
\subsection{Key Modules of CE2P} 
Our CE2P integrates the local fine details, global context and semantic edge context into a unified network. The overview of our framework is shown in Fig.~\ref{fig:framework}. Specifically, it consists of three key components to learn for parsing in an end-to-end manner, \ie context embedding module, high-resolution embedding module and edge perceiving module. ResNet-101 is adopted as the feature extraction backbone. 

\noindent \textbf{Context Embedding Module}
Global context information is useful to distinguish the fine-grained categories. For instance, the left and right shoe have relatively high similarity in appearance. To differentiate between the left and right shoe, the global information, like the orientation of leg and body, provides an effective context prior. As we know, feature pyramid is a powerful way to capture the context information. Draw on the previous work PSP~\cite{Zhao2017}, we utilize a pyramid pooling module to incorporate the global representations. We perform four average pooling operations on the features extracted from ResNet-101 to generate multiple scales of context features with size 1$\times$1, 2$\times$2, 3$\times$3, 6$\times$6 respectively. Those context features are upsampled to keep the same size with the original feature map by bilinear interpolation, which are further concatenated with the original feature. Then, the 1$\times$1 convolution is employed to reduce the channels and better integrate the multi-scale context information. Finally, the output of context embedding module is fed into the following high-resolution module as global context prior.  

\noindent \textbf{High-resolution Embedding Module}
In human parsing, there exist several small objects to be segmented, \eg socks, shoes, sunglasses and glove. Hence, high-resolution feature for final pixel-level classification is essential to generate an accurate prediction. To recover the lost details, we adopt a simple yet effective method which embeds the low-level visual features from intermediate layers as complementary to the high-level semantic features. We exploit the feature from the \emph{conv2} to capture the high-resolution details. The global context feature is upsampled by factor 4 with bilinear interpolation, and concatenated with local feature after channel reduced by 1$\times$1 convolution. Finally, we conduct two sequential 1$\times$1 convolution on the concatenated feature to better fuse the local and global context feature. In this manner, the output of high-resolution module simultaneously acquires high-level semantic and high-resolution spacial information.       

\noindent \textbf{Edge Perceiving Module}
This module aims at learning the representation of contour to further sharp and refine the prediction. We introduce three branches to detect multi-scale semantic edges. As illustrated in Fig.~\ref{fig:framework}, a 1$\times$1 convolution are conducted to \emph{conv2}, \emph{conv3} and \emph{conv4} to generate 2-channel score map for the semantic edge. And then, 1$\times$1 convolution is performed to obtain the fused edge map. Those intermediate features of edge branches, which can capture useful characteristics of object boundaries, are upsampled and concatenated with the features from high-resolution. Finally, 1$\times$1 convolution is performed on the concatenated feature map to predict the pixel-level human parts.      

Our CE2P consisting of the three modules is learned with an end-to-end manner. The outputs of CE2P consist of two parsing results and edge prediction. Hence, the loss can be formulated as:

\begin{equation}
  L = L_{Parsing} + L_{Edge} + L_{Edge-Parsing} ,
\end{equation}
where $L_{Edge}$ denotes the weighted cross entropy loss function between the edge map detected by edge module and the binary edge label map; $L_{Parsing}$ denotes the cross entropy loss function between the parsing result from high resolution module and the parsing label; $L_{Edge-Parsing}$ denotes the cross entropy loss function between the final parsing result, which is predicted from the edge perceiving branch, and the parsing label.

% where $L_{Parsing}$ and $L_{Edge-Parsing}$ are the loss functions for human parsing from high resolution and edge perceiving module, respectively. $L_{Edge}$ is the loss function for edge prediction from the edge perceiving module. 

\subsection{Multiple Human Parsing (MHP)}
\begin{figure}[t]
    \begin{center}
        \includegraphics[width=1.0\linewidth]{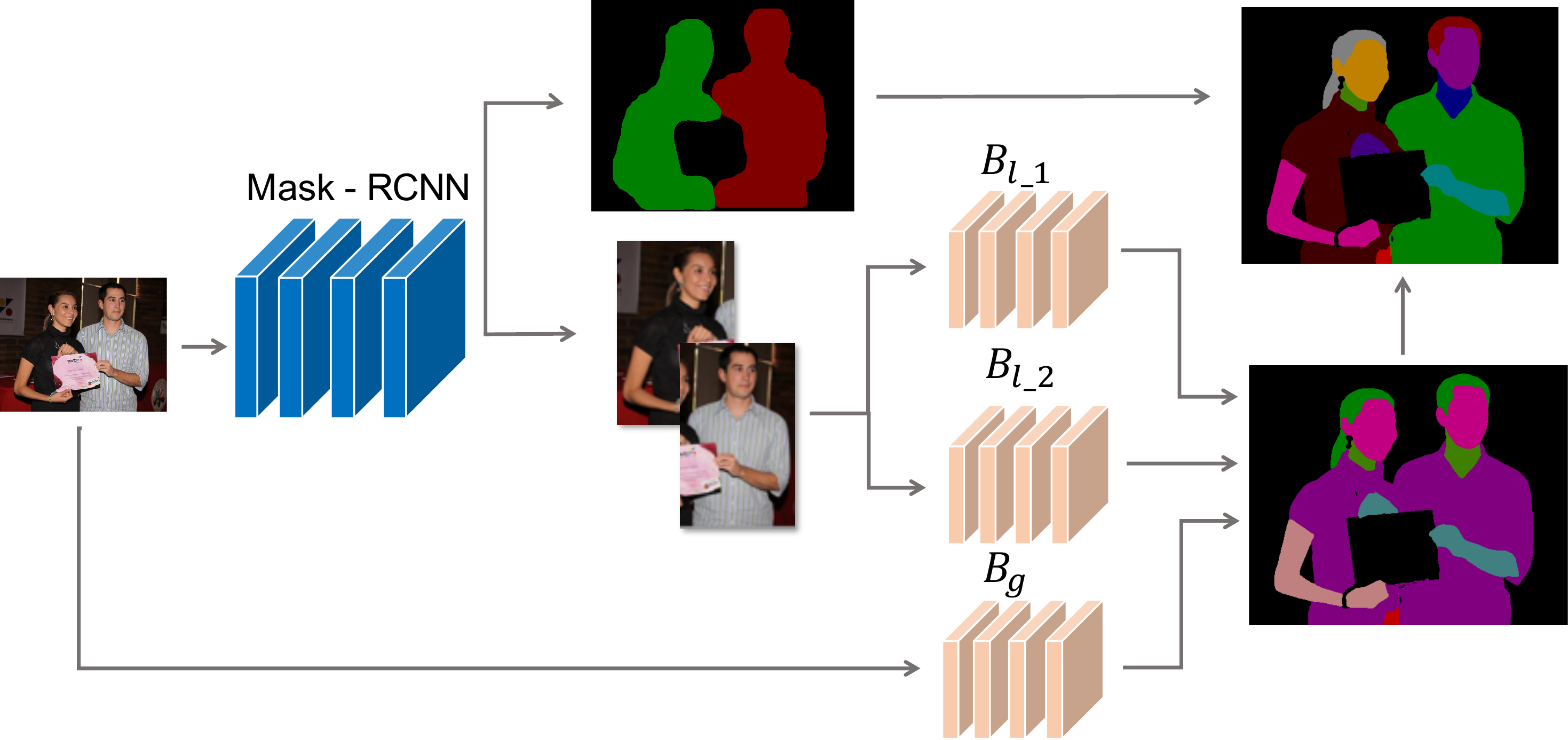}
    \end{center}
    \caption{The overall framework of CE2P for Multiple Human Parsing task.}
    \label{fig:mhp_overall}
\end{figure}

MHP is a more challenging task, which not only needs to classify the semantics of pixels but also identify the instance (\ie one unique person) that these pixels belong to. To achieve high-quality parsing results in the scenario of multiple persons, we design a framework called M-CE2P upon our CE2P and Mask R-CNN~\cite{he2017mask}. As shown in Fig.~\ref{fig:mhp_overall}, our M-CE2P leverages three branches, denoted by $B_g$, $B_{l\_1}$, $B_{l\_2}$, to make predictions from global to local views. The details of the three branches are introduced in the following.

\noindent \textbf{Global Parsing $B_g$} 
In spite of the CE2P is proposed for single human parsing, we find it shows considerable performance on multiple human images as well. Hence, we first apply it over the entire image for global parsing. For the branch of $B_g$, we train a CE2P model with the entire images. Then, the output of this branch is leveraged as complementary to the following local parsing. The global parsing branch can provide context information when there exist occlusions among multiple persons. For instance, the same semantic parts form different persons can be easy to tell apart, and the spatial relationship among persons can be captured to handle the circumstance of occlusion. However, it does not concentrate on the relatively small human instances. As a result, body parts belonging to small-scale person are likely ignored by $B_g$.   
%Similar to single human parsing task, the global parsing branch($B_g$) also aims at segmenting human body parts that an image contains. For $B_g$, we using raw data to train the CE2P, and directly exploit it to perform the forward propagation. With this one-stage manner, we can obtain an instance-agnostic parsing result associated with the input image.

\noindent \textbf{Local Parsing with Predicted Instance Masks $B_{l\_1}$}
 To alleviate the problem of global parsing $B_g$, we consider locating the persons as a preprocessing step to generate accurate parsing results. Towards this end, we propose a two-stage branch devoting to human-level local parsing. Specifically, we employ Mask R-CNN~\cite{he2017mask} to extract all the person patches in the input image, and resize them to fit the input size of CE2P. And then, all the human-level sub-images are fed into CE2P to training the model for local view. During inference stage, the sub-images with single human instance of input images are extracted by Mask R-CNN, and further fed into the trained model to make parsing predictions. The predicted confidence maps are resized to the original size by bilinear interpolation for the following prediction over the entire image. Finally, the confidence maps for each sub-images are padded with zeros to keep the same size as the confidence map from $B_g$, and further fused together by element-wise summation on foreground channels and minimization on background channel. 

\noindent \textbf{Local Parsing with Ground-truth Instance Masks $B_{l\_2}$} Considering that a human instance obtained from the ground-truth instance mask is more approximate to the real single human image, we introduce the branch $B_{l\_2}$ to train a model with the data generated from ground-truth instance mask. This branch is rather similar with $B_{l\_1}$, the only difference is that we obtain person patches with the guide of the ground truth bounding boxes in the training stage. With the $B_{l\_2}$, the performance of local parsing can be further boosted.  Finally, the predictions generated by the three branches are fused by element-wise summation to obtain the final instance-agnostic parsing result. The predicted instance-agnostic parsing results are further fed into the following process to make the instance-level parsing.  

\noindent \textbf{Instance-level Parsing and Label Refinement}
With the instance-agnostic parsing result obtained from M-CE2P, we consider two aspects to generate the instance-level label, \ie \emph{instance assignment} for predicting instance-aware results and \emph{label refinement} for solving the shortage of the under-segmentation phenomenon of Mask R-CNN. For instance assignment, we directly apply human masks generated by Mask R-CNN to assign instance-level part label of global body parts. Concretely, parts will be assigned with different part instance labels when they are same category while belonging to different masks. Through the experiments, we find that the parsing mask predicted from our CE2P is more reliable than human instance map. To further validate the reliability of parsing results, we introduce the label refinement by expanding the area of intersection with the neighbor pixels which have same parsing label while exceeding the human instance. For example, some regions of marginal parts(\eg hair, hands) are very likely to be outside of the area of predicted human masks. We use searching based method to alleviate this problem. Specifically, for each border pixel of the part obtained from the assignment step, we use Breadth-First Search to find the pixel that is endowed with an instance-agnostic class label but no part label due to the inaccuracy of segmentation prediction. With the proposed refinement, the body parts excluded by human mask can be effectively included in the final instance-level result.

\section{Experimental Results}

\begin{table*}[htbp]
    \caption{Comparison of CE2P in various module settings on the validation set of LIP. The results are obtained without left-right flipping except for the last row. 'B' means baseline model. 'G', 'H' and 'E' denote global context, high resolution and edge perceiving module, respectively.  }
    \label{tab:IoU}
    \centering
    %\begin{tabular}{lccccccccccccccccccccc}
    \resizebox*{!}{3.2cm}{
    \begin{tabular}{llllllllllllllllllllll}
    \toprule 
        Method &  \rotatebox{90}{bkg} & \rotatebox{90}{hat} & \rotatebox{90}{hair} & \rotatebox{90}{glove} & \rotatebox{90}{glasses} & \rotatebox{90}{u-cloth} & \rotatebox{90}{dress} & \rotatebox{90}{coat} & \rotatebox{90}{socks} & \rotatebox{90}{pants} & \rotatebox{90}{j-suits} & \rotatebox{90}{scarf} & \rotatebox{90}{skirt} & \rotatebox{90}{face} & \rotatebox{90}{l-arm} & \rotatebox{90}{r-arm} & \rotatebox{90}{l-leg} & \rotatebox{90}{r-leg} & \rotatebox{90}{l-shoe} & \rotatebox{90}{r-shoe} & \rotatebox{90}{mIoU}  \\
        \midrule 
B & 86.08 & 63.24 & 70.14 & 33.75 & 29.45 & 66.15 & 24.10 & 51.79 & 45.70 & 70.75 & 21.52 & 13.56 & 20.36 & 73.04 & 58.24 & 60.88 & 49.70 & 48.06 & 36.50 & 36.40 & 47.97 \\ 
B+H & 86.69 & 64.65 & 71.53 & 33.76 & 33.30 & 66.50 & 25.11 & 51.52 & 46.10 & 71.46 & 23.62 & 14.77 & 21.44 & 74.08 & 60.04 & 62.41 & 50.36 & 50.74 & 36.95 & 37.19
 & 49.11  \\   
B+G & 86.61 & 63.77 & 71.01 & 34.90 & 30.73 & 67.93 & 30.89 & 54.14 & 46.11 & 72.21 & 23.38 & 13.16 & 20.16 & 73.50 & 59.51 & 62.12 & 50.70 & 50.48 & 38.63 & 38.84 & 49.44  \\ 
B+E & 86.87 & 63.47 & 71.56 & 35.60 & 31.35 & 67.06 & 27.80 & 52.28 & 47.85 & 72.21 & 23.89  & 15.60 & 20.57 & 74.37 & 61.04 & 63.36 & 53.30 & 52.70 & 39.72 & 40.16 & 50.04  \\
B+G+H & 87.22 & 65.34 & 72.13 & 36.18 & 31.97 & 68.86 & 31.02 & 55.81 & 47.35 & 73.23 & 26.91 & 12.28 & 20.58 & 74.49 & 62.95 & 65.18 & 56.31 & 55.59 & 43.49 & 43.80 & 51.54 \\  
\tabincell{c}{B+G+H+E\\(CE2P)} & 87.41 & 64.62 & 72.07 & 38.36 & 32.20 & 68.92 & 32.15 & 55.61 & 48.75 & 73.54 & 27.24 & 13.84 & 22.69 & 74.91 & 64.00 & 65.87 & 59.66 & 58.02 & 45.70 & 45.63 & 52.56 \\  
\tabincell{c}{CE2P\\(Flipping)} & \bf{87.67} & 65.29 & \bf{72.54} & \bf{39.09} & \bf{32.73} & \bf{69.46} & \bf{32.52} & \bf{56.28} & \bf{49.67} & \bf{74.11} & 27.23 & \bf{14.19} & 22.51 & \bf{75.50} & \bf{65.14} & \bf{66.59} & \bf{60.10} & \bf{58.59} & \bf{46.63} & \bf{46.12} & \bf{53.10} \\ 
        \bottomrule 
    \end{tabular}
    }
\end{table*}   

\subsection{Dataset and Metrics}
We compare the performance of single human parsing of our proposed approach with other state-of-the-arts on the LIP~\cite{Liang2018} dataset, and we further evaluate the multiple human parsing on CIHP~\cite{gong2018instance} and MHP v2.0~\cite{Li2017} dataset.

\noindent\textbf{LIP dataset}: The LIP~\cite{Liang2017} dataset is used in LIP challenge 2016, which is a large-scale dataset that focuses on single human parsing. There are 50,462 images with fine-grained annotations at pixel-level with 19 semantic human part labels and one background label. Those images are further divided into 30K/10K/10K for training, validation and testing, respectively.

\noindent\textbf{CIHP dataset}: CIHP~\cite{gong2018instance} provides a dataset with 38,280 diverse human images, in which contains 28,280 training, 5K validation and 5K test images. The images have pixel-wise annotation on 20 categories and instance-level identification.   

\noindent\textbf{MHP v2.0 dataset}: The MHP v2.0 dataset is designed for multi-human parsing in the wild including 25,403 images with finer categories up to 58 semantic labels. The validation set and test set have 5K images respectively. The rest 15,403 are provided as the training set.  

\noindent\textbf{Metrics}: We use mean IoU to evaluate the global-level predictions, and use the following three metrics to evaluate the instance-level predictions. $Mean~AP^r$ computes the area under the precision-recall curve with the limitation of a set of IoU threshold, and figure out the final averaging result, which is first introduced in~\cite{hariharan2014simultaneous}; $AP^p$~\cite{Li2017} computes the pixel-level IoU of semantic part categories within a person, instead of global circumstance; $PCP$~\cite{Li2017} elaborates how many body parts are correctly predicted of a certain person, guided with pixel-level IoU.

\subsection{Implement Details}

  We implement the proposed framework in PyTorch~\cite{paszke2017automatic} based on~\cite{torch2018segment}, and adopt ResNet-101~\cite{He2016} as the backbone network. The input size of the image is 473$\times$473 during training and testing. We adopt the similar training strategies with Deeplab~\cite{Chen2018a}, \ie ``Poly'' learning rate policy with base learning rate 0.007. We fine-tune the networks for approximately 150 epochs. For data augmentation, we apply the random scaling (from 0.5 to 1.5), cropping and left-right flipping during training. Note that the edge annotation used in the edge perceiving module is directly generated from the parsing annotation by extracting border between different semantics.

\begin{table}[htbp]
    \footnotesize
    \caption{Comparison of performance on the validation set of LIP with state-of-arts methods.}
    \label{tab:IoU_others}
    \centering
    \begin{tabular}{llll}
        \toprule 
        Method & pixel acc. & mean acc. & mIoU  \\ \midrule
        DeepLab (VGG-16) & 82.66 & 51.64 & 41.64 \\  
        Attention & 83.43 & 54.39 & 42.92 \\  
        DeepLab (ResNet-101) & 84.09 & 55.62 & 44.80 \\  
        JPPNet~\cite{Liang2018} &  86.39 & 62.32 &  51.37 \\  
        CE2P & \bf{87.37} & \bf{63.20} & \bf{53.10} \\ \bottomrule 
    \end{tabular} 
\end{table}

\subsection{Single Human Parsing} 
\begin{figure}[t]
   \centering
     \includegraphics[width=0.46\textwidth]{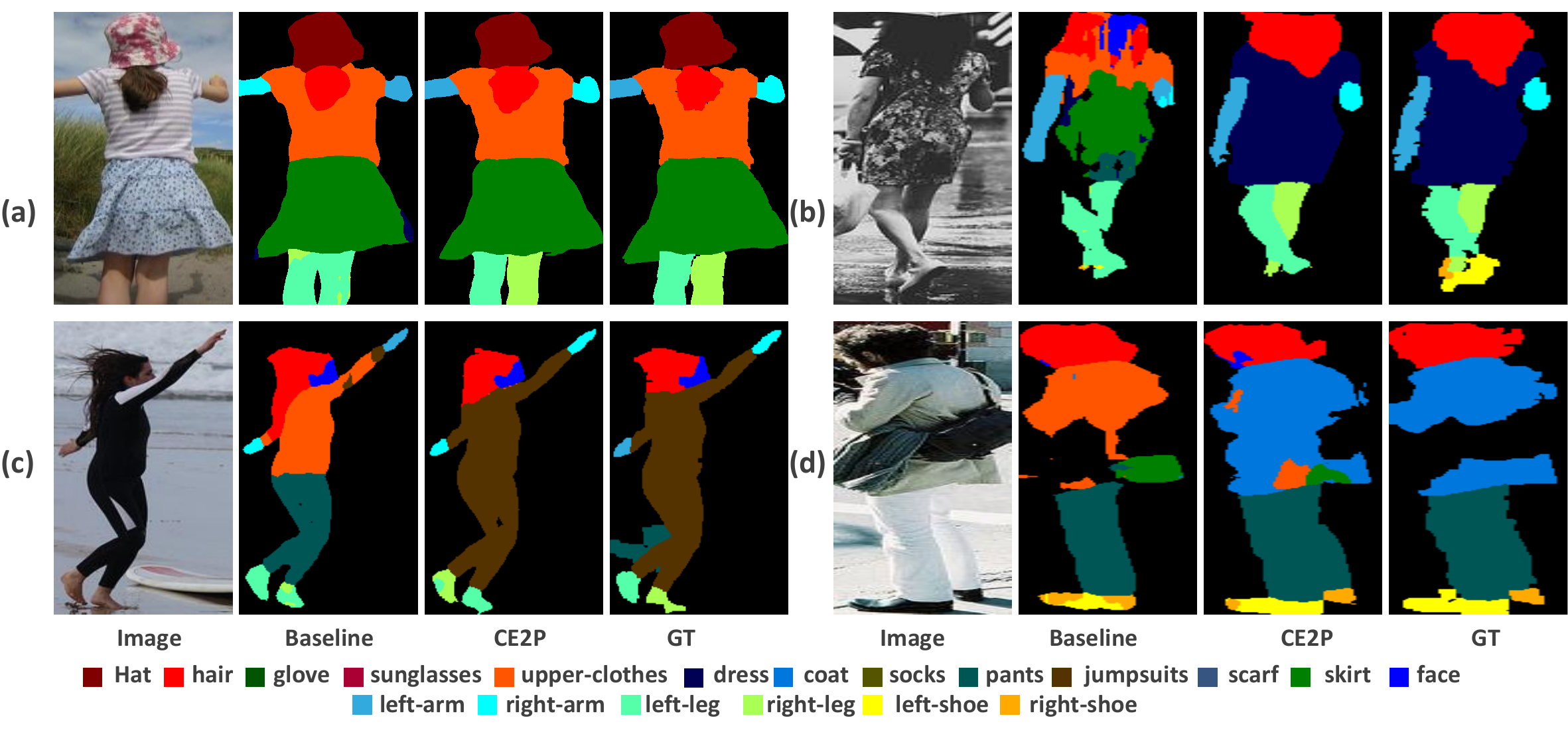}
     \caption{Some issues of baseline model on LIP dataset. (a) The left arm (leg) is wrongly predicted as right arm (leg). (b) The dress is separated into a upper-clothes and skirt. (c) The jumpsuits is parsed into a upper-clothes and pants. (d) The coat is mislabeled as upper-clothes.   }
     \label{fig:comparision}
\end{figure}

To investigate the effectiveness of each module, we report the performance under several variants of CE2P in Tab.~\ref{tab:IoU}. We begin the experiment with a baseline model without any proposed modules. For our baseline, the prediction is directly performed on the final feature map extracted from ResNet-101. The resolution of the final feature map is $1/16$ to the input size. The results are shown in Tab.~\ref{tab:IoU}, and we can see the baseline model reaches 47.97\% accuracy. Some failure examples are shown in Fig.~\ref{fig:comparision}. Observing the per-classes performance and the visualized results, there exist the following problems.  1) Big-size objects have the discontinuous prediction. For instance, the dress is always parsed as a upper-clothes and skirt, and the jumpsuit is separated into a upper-clothes and pants. 2) The confusable categories are hard to distinguish. \ie the coat and upper-clothes. 3) The left and right parts are easily confused, which frequently occurs in the back-view human body and the front-view body with legs crossed.

\noindent\textbf{Global Context Embedding Module} To evaluate the effectiveness of each module, we first conduct experiments by introducing a global context embedding module. In our architecture, we leverage the pyramid pooling~\cite{Zhao2017} to generate multi-scale context information. As shown in Tab.~\ref{tab:IoU}, we can find it brings about 1.5\% improvements on mIoU, which demonstrates that the multi-scale context information can assist the fine-grained parsing. Particularly, it shows significant boosts (nearly 7\%) in the class of dress. Since the long-range context information can provide the more discriminated characteristic, the global context embedding module is helpful for the big-size objects.   
   
\noindent\textbf{High Resolution Embedding Module} To figure out the importance of the high resolution, we conduct experiments by further introducing a high resolution module. From Tab.~\ref{tab:IoU}, we can find that the performance gains nearly 2\% improvement with the high resolution embedding module. As human parsing is a fine-grained pixel-wise classification, it requires a lot of detailed information to identify the various small-size parts accurately. With high resolution embedding module, the features from shallow and high-resolution layers provide more details not available in deep layers. The results demonstrate the effectiveness as well.    

\noindent\textbf{Edge Perceiving Module} Finally, we report the performance with the edge perceiving module in Tab.~\ref{tab:IoU}. Based on the above two modules, appending edge perceiving module still brings nearly 1\% boosts. That's the contours of the parts can be underlying constraints during separating the semantic parts from a human body. In addition, the features from multiple edge branches carrying various details of the objects can further promote the human parts prediction. Finally, fusing with the flipped images gives 0.6\% gain. 

\noindent \textbf{Comparison with State-of-the-Arts} 
We evaluate the performance of CE2P on the validation dataset of LIP and compare it to other state-of-the-art approaches. The results are reported in Tab.~\ref{tab:IoU_others}. First, we note that the mIoU of our CE2P significantly outperforms other approaches. The improvement over the state-of-the-art method validates the effectiveness of our CE2P for human parsing. When comparing with the current state-of-the-art approach JPPNet, our method exceeds by 1.73\% in terms of mIoU. In particular, the performance on the small-size categories, \ie `socks' and `sunglasses', yields obviously improvement. Thanks to the high resolution embedding and edge perceiving module, the details and characteristic of small objects can be captured for further pixel-level classification.  Besides, the JPPNet achieves the performance of 51.37\% by utilizing extra pose annotation with a complex network structure. Nevertheless, our CE2P obtains better performance with a simpler network structure and no need for extra annotation.
\begin{table}[htbp]
    \footnotesize
    \caption{Comparison of the performance on the test set of single and multiple human parsing datasets. Here the subscript $m$ in $X_{m}$ means the mean value of $X$.}
    \label{tab:singleChallenge}
    \centering
    %\resizebox{!}{3.2cm}{
    \begin{tabular}{lccc}
    \toprule         
    Method & pixel acc. & mean acc.  & mIoU  \\ \midrule 
    \multicolumn{4}{l}{Single human parsing (Track 1)} \\  
     \bf{ours} & \bf{88.92} & \bf{67.78} & \bf{57.90}  \\   
     \bf{ours (single model)} & 88.24 &  67.29 &  56.50 \\  
    JD\_BUPT & 87.42 & 65.86 & 54.44  \\    
    AttEdgeNet & 87.40 & 67.17 & 54.17 \\   
    \midrule        
    Method & mIoU & $AP^r_{m}$ & $AP^r_{0.5}$   \\\midrule
    \multicolumn{4}{l}{Multiple human parsing on CIHP (Track 2)} \\
    \bf{ours}              &  \bf{63.77} & \bf{45.31}  & \bf{50.94}     \\    
    DMNet(2nd place)            &  61.51      & 41.50       & 46.12          \\
    PGN~\cite{gong2018instance} & 55.80      & 33.60       & 35.80          \\\midrule        
    Method               & $PCP_{0.5}$ & $AP^p_{0.5}$ & $AP^p_{m}$   \\\midrule 
    \multicolumn{4}{l}{Multiple human parsing on MHP v2.0 (Track 5)} \\
    \bf{ours}   & \bf{41.82} & \bf{33.34}  & \bf{42.25}     \\
    S-LAB(2nd place)     & 38.27      & 31.47       & 40.71          \\
    NAN~\cite{Zhao2018}  & 32.25      & 25.14       & 41.78          \\
    \bottomrule  
    \end{tabular}
    %}
\end{table}

\begin{table*}[t]
    \footnotesize
    \begin{center}
        \caption{Comparisons on MHP v2.0 validation dataset} \label{tab:res_on_mhpv2}        
        \begin{tabular}{lccccccc}
            \toprule
            Method                               & mIoU        & $AP^r_{0.5}$    & Mean $AP^r$     & $AP^p_{0.5}$    & Mean $AP^p$     & $PCP_{0.5}$     & Mean $PCP$      \\
            \midrule
            NAN~\cite{Zhao2018}                    & -               & -               & -               & 24.83           & 42.77           & 34.37           & -               \\
            \bottomrule
            $B_g$                             & 38.25          & 27.32          & 24.96          & 21.63          & 37.13          & 32.59          & 34.94          \\
            $B_{l\_1}$                          & 40.18          & 30.82          & 27.95          & 24.98          & 38.72          & 36.42          & 37.54          \\
            $B_{l\_2}$                            & 40.30          & 30.75          & 28.04          & 24.46          & 38.60          & 36.09          & 37.46          \\

            $B_{l\_1} + B_g$                        & 40.48          & 30.63          & 27.79          & 29.62          & 40.52          & 39.29          & 38.46          \\
            $B_{l\_2} + B_g$                          & 40.52          & 30.67          & 27.89          & 29.29          & 40.56          & 39.13          & 38.50          \\
            $B_{l\_1} + B_{l\_2}$                       & 41.05          & 31.63          & 28.82          & 28.71          & 40.37          & 39.32          & 38.90          \\
            $B_{l\_1} + B_{l\_2} + B_g$(M-CE2P)                     & 41.11          & 31.50          & 28.60          & 30.92          & 41.29          & 40.58          & 39.32          \\
            \midrule
            $B_g$ with refinement                 & 38.25          & 29.70          & 26.94          & 24.04          & 38.21          & 35.09          & 36.36          \\
            $B_{l\_1}$ with refinement              & 40.18          & 33.69          & 30.34          & 28.20          & 40.03          & 39.56          & 39.21          \\
            $B_{l\_2}$ with refinement                & 40.30          & 33.66          & 30.41          & 27.43          & 39.84          & 39.09          & 39.11          \\

            $B_{l\_1} + B_g$ with refinement            & 40.48          & 33.39          & 30.08          & 32.84          & 41.90          & 42.32          & 40.19          \\
            $B_{l\_2} + B_g$ with refinement              & 40.52          & 33.54          & 30.22          & 32.62          & 41.88          & 42.14          & 40.19          \\
            $B_{l\_1} + B_{l\_2}$ with refinement           & 41.05          & \textbf{34.58} & \textbf{31.24} & 32.39          & 41.75          & 42.64          & 40.66          \\
            $B_{l\_1} + B_{l\_2} + B_g$(M-CE2P) with refinement         & \textbf{41.11} & 34.40          & 30.97          & \textbf{34.47} & \textbf{42.70} & \textbf{43.77} & \textbf{41.06} \\ 
            %$M_{mask+gt+w}$ with refinement on CIHP & 63.59          & 54.66          & 48.92          & 61.73          & 51.74          & 59.35          & 49.54          \\
            \bottomrule
        \end{tabular} 
    \end{center}
\end{table*}

\subsection{Multiple Human Parsing}
We provide abundant experimental results on two large datasets of multiple human parsing task, named CIHP and MHP v2.0. 
%As what we expected, the result generated by three-branch fusion strategy with refinement can attach the best performance under more trustworthy metrics.
The results are exhibited in %Table~\ref{tab:res_on_lip} and
Tab.~\ref{tab:res_on_mhpv2}. Due to space limitation, we only show detailed results on the more challenging MHP v2.0 dataset. The detailed explanation will be given in the following paragraphs.

\noindent \textbf{Label Refinement}
A remarkable improvement attributes the success to the label refinement operation. As %Table~\ref{tab:res_on_lip} and
Tab.~\ref{tab:res_on_mhpv2} shows, it exactly brings performance boosting, despite of the combination strategies. As we mentioned before, the results provided by Mask R-CNN are likely to ignore some partial area of marginal body parts, especially on complex images. However, CE2P may catch these parts from the localized human sub images. Therefore, the refinement operation can alleviate the under segmentation problem. For clarity, the following analyses are all based on refined results.

\begin{figure}[t]
    \begin{center}
        \includegraphics[width=0.8\linewidth]{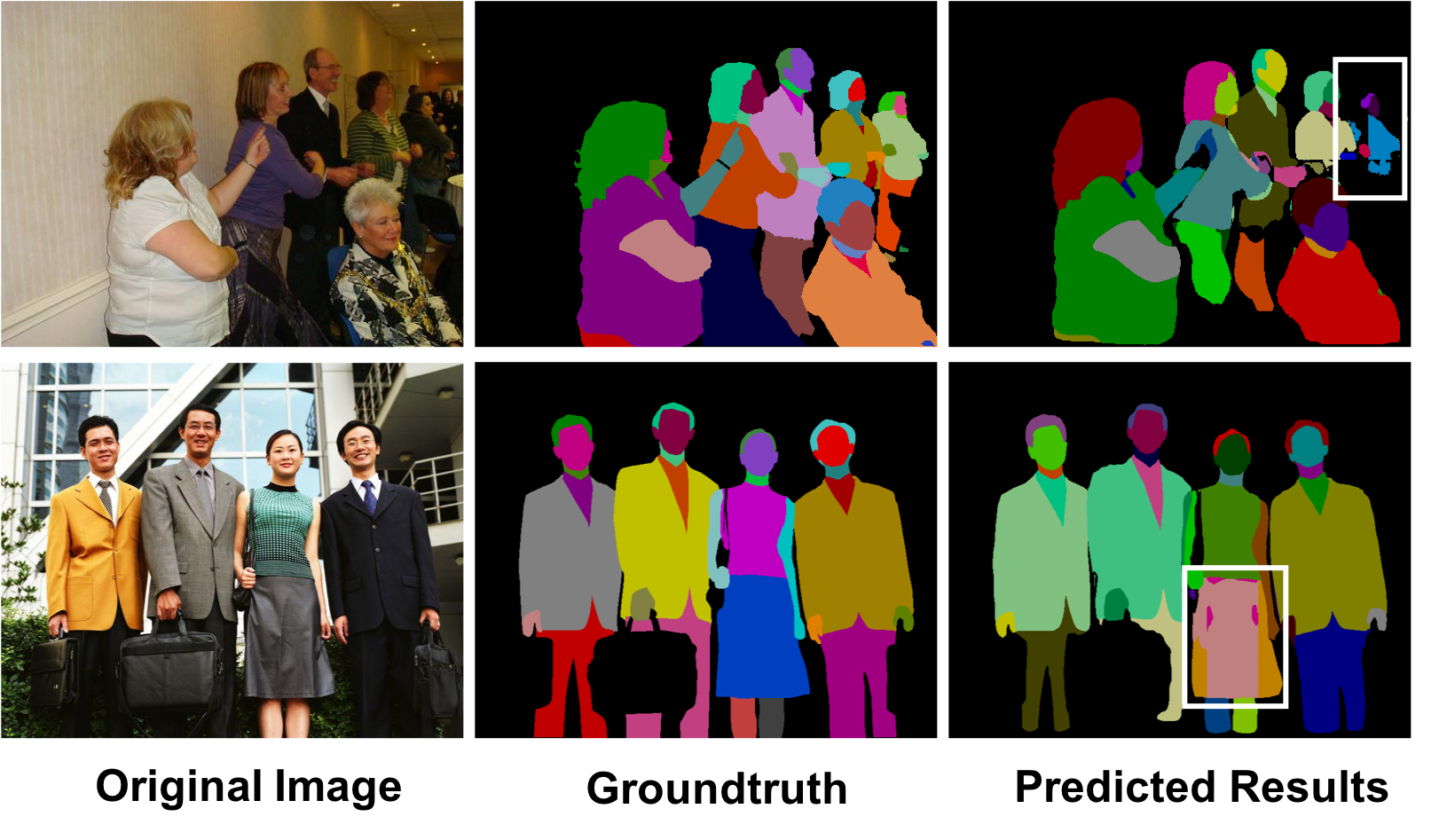}
    \end{center}
    \caption{A visualized example for Multiple Human Parsing task. The areas surrounded by white bounding boxes indicate the failure cases of predicted results.}
    \label{fig:mhp_expe}  
\end{figure}

\noindent \textbf{Comparison with Various Combination Strategies}
To prove the effectiveness of the multi-branch fusion strategies, we perform experiments with various branch combinations. From Tab.~\ref{tab:res_on_mhpv2}, we can notice that the results produced by the double-branch model are better than that produced by the single-branch model, and the all-branch model, \ie M-CE2P, catch the best performance on most of the metrics. Especially on more convincing human-centric metrics like $AP^p$ and $PCP$, M-CE2P shows out a significant performance boosting than all the other models. It proves that branches can make complements with each other. As mentioned in the previous section, the branch $B_g$ trained with the entire image lacks the ability to grab small-scale persons in a scene, such as the first image in Fig.~\ref{fig:mhp_expe}. Hence, it only achieves the performance of 24.04\% in terms of $AP^p_{0.5}$. Nevertheless, this shortcoming can be compensated by $B_{l\_1}$ and $B_{l\_2}$ to capture a more precise view of local context. On the other hand, the benefits from $B_g$ are still cannot be ignored. %It can reduce the deviation caused by the human-level fusion of the other two models, and better handling the border area segmentation between different persons.
The global context that $B_g$ has makes a performance improvement of 2.08\% in terms of $AP^p_{0.5}$ than the result only utilizes $B_{l\_1}$ and $B_{l\_2}$. Finally, the all-branch fused M-CE2P can reach the best performance under the $AP^p$ and $PCP$ metrics.
%Finally, it can reach the best performance under $AP^p$ metric, which is more reliable on multiple human parsing task than $AP^r$.

\noindent \textbf{Comparison with State-of-the-Arts}
Compared to other state-of-the-art methods, our M-CE2P model still maintains a large-margin leading edge on major metrics, as Tab.~\ref{tab:singleChallenge} and Tab.~\ref{tab:res_on_mhpv2} illustrate. On validation set of MHP v2.0, our proposed model outperforms than~\cite{Zhao2018} 9.64\%, 9.40\% in terms of $AP^p_{0.5}$ and $PCP_{0.5}$, respectively;%, and comparable in terms of mean $AP^p$, which is only under-perform 0.07\%;
Furthermore, on the test set of MHP v2.0, we outperform than~\cite{Zhao2018} 8.20\%, 9.57\% and 0.47\% in terms of $AP^p_{0.5}$, $PCP_{0.5}$ and mean $AP^p$, respectively; On the test set of CIHP, our M-CE2P performs 63.77\%, 45.31\%, 50.94\% in terms of mIoU, mean $AP^r$ and $AP^r_{0.5}$, respectively, which all outperform than~\cite{gong2018instance}. Above all, we investigate and combine several useful properties to improve the parsing results. Each module plays an important role and makes improvements for the final results. Benefiting from our parsing network, the prediction on the whole multi-person image already has a comparable result with baseline methods. Based on the accurate parsing results, our fusion strategy and label refinement can make further boost.
  
\noindent \textbf{Visualization}
Some visualized results with their failure cases are shown in Fig.~\ref{fig:mhp_expe}. Generally speaking, our M-CE2P can handle a rather complex scene with multiple persons, and produce a satisfactory result. %Thanks to local parsing, most small parts, such as hands and parts of small-scale people, can be clearly segmented; And owe to the global context that $B_g$ offered, most occlusion can also be correctly handled.
However, there are also some failure cases produced by our M-CE2P. From the perspective of fusion strategy, it has following problems: 1) Under some circumstances, the $B_g$ brings too much negative confidence to parts belong to human far away from camera, so that the confidence provided by local parsing may be drastically reduced; 2) When acting sub-image fusion, the local information of tightly closed humans may be disturbed with each other. For example, in bottom line of Fig.~\ref{fig:mhp_expe}, the man with black suit make the edge of woman's skirt strongly be mistaken as other semantics.

\subsection{Results in Challenge}

With our CE2P framework, we achieved the 1st places of all three human parsing tracks in the 2nd Look into Person (LIP) Challenge. Tab.~\ref{tab:singleChallenge} shows a few of the results on the single human parsing lead-board. By integrating the results from three models with different input size, our CE2P achieved 57.9\%. More importantly, our single model already attained the state-of-arts performance without any bells and whistles. Besides, we design a M-CE2P upon our CE2P for multiple human parsing with three branches to predict from global to local view. Benefiting from the M-CE2P model, we achieved high performance in all the multiple human parsing benchmarks without refinement process, \ie CIHP and MHP v2.0, respectively. Specifically, we yielded 45.31\% of Mean $AP^r$ and 33.34\% of $AP^p_{0.5}$, which outperform the second place more than 3.81\% 1.87\%, respectively.  

\section{Conclusion}

In this paper, we investigate several useful properties for human parsing, including feature resolution, global context information and edge details. We design a simple yet effective CE2P system, which consists of three key modules to incorporate the context and detailed information into a unified framework. Specifically, we use a high-resolution embedding module to capture details from shallow layer, a global context embedding module for multi-scale context information, and an edge perceiving module to constrain object contours. For multiple human parsing, we fuse three CE2P based model to generate global parsing prediction, and use a straight-forward way to produce the instance-level result with the guide of human mask. The experimental results demonstrate the superiority of the proposed CE2P.

\section{Acknowledgment}

%This work was supported in part by Fundamental Research Funds for the Central Universities (No. 2017YJS048, No. 2018JBZ001), National Key Research and Development of China (No.2016YFB0800404, 2017YFC1703503), National Natural Science Foundation of China (No.61532005, No.61572065), Joint Fund of Ministry of Education of China and China Mobile (No.MCM20160102), IBM-ILLINOIS Center for Cognitive Computing Systems Research (C3SR) - a research collaboration as part of the IBM AI Horizons Network.

This work was supported in part by National Key Research and Development of China (No.2017YFC1703503), National Natural Science Foundation of China (No.61532005, No.61572065),  Fundamental Research Funds for the Central Universities (No. 2017YJS048, No. 2018JBZ001), Joint Fund of Ministry of Education of China and China Mobile (No.MCM20160102), IBM-ILLINOIS Center for Cognitive Computing Systems Research (C3SR) - a research collaboration as part of the IBM AI Horizons Network.
 
 \bibliographystyle{aaai}
 \bibliography{aaai}

\end{document}